\documentclass[letterpaper, 10 pt, conference]{ieeeconf}
\IEEEoverridecommandlockouts
\usepackage{fancyhdr}
\fancypagestyle{preprint}{
  \fancyhf{}
  \fancyhead[L]{PREPRINT VERSION}
  
  \setlength{\topmargin}{-0.7in}
  \setlength{\headheight}{15pt}
  \setlength{\headsep}{25pt}
}

\usepackage{amsmath}
\usepackage{amssymb}
\usepackage{booktabs}
\usepackage{subcaption}
\usepackage{graphicx}
\usepackage{multirow}
\usepackage{array}
\usepackage[hyphens]{url}
\usepackage{cite}
\newcommand{\algoname}{FEEL}
\newcommand{\algonamelong}{\textbf{F}inger \textbf{E}xternal-Force \textbf{E}stimation from Proprioceptive Signa\textbf{L}s}

\title{Proprioceptive Finger Force Estimation for Reliable Box Manipulation}
\author{Junha Cha$^{1\dagger}$, Chanwoo Kim$^{1\dagger}$, Kisang Park$^{1}$, Yeoncheol Jang$^{1}$, Doyoung Kim$^{1}$,\\ Seungmin Kim$^{1}$, Kyungjae Lee$^{2}$, and Sungjoon Choi$^{1,3}\textsuperscript{*}$%
\thanks{$^{\dagger}$Junha Cha and Chanwoo Kim contributed equally to this work and share first authorship.}
\thanks{$^{1}$Junha Cha, Chanwoo Kim, Kisang Park, Yeoncheol Jang, Doyoung Kim, Seungmin Kim, and Sungjoon Choi are with the Department of Artificial Intelligence, Korea University, Seoul, Republic of Korea. (e-mails: chajunha1211@gmail.com; \{chanwoo-kim, qkrrltkd1626, keep9642, ehdud1684, seungmin-kim, sungjoon-choi\}@korea.ac.kr)}
\thanks{$^{2}$Kyungjae Lee is with the Department of Statistics, Korea University, Seoul, Republic of Korea. (e-mail: kyungjae\_lee@korea.ac.kr)}
\thanks{$^{3}$Sungjoon Choi is with RLWRLD, Seoul, Republic of Korea.}%
}

\begin{document}
\maketitle
\thispagestyle{preprint}
\pagestyle{preprint}

\begin{abstract}
Reliable box manipulation requires successful placement without damage, which depends on maintaining sufficient contact force without excessive loading.
We hypothesize that per-finger force feedback improves manipulation reliability relative to open-loop finger commands.
To this end, we present \algonamelong{} (\algoname), an uncertainty-aware residual method that estimates per-finger external force magnitude from proprioceptive histories.
Our method combines a calibrated virtual-spring estimate with a residual distribution predicted by a Mixture Density Network (MDN) and requires no dedicated contact sensors during deployment.
The estimated force is used for continuous per-finger adaptation, while predictive uncertainty is evaluated as an empirical signal of ambiguity in proprioceptive force estimation.
Across ten fingers and contact forces spanning 0--20~N, our method achieves a mean RMSE of 0.251~N, reducing the error by 92.8\% relative to the virtual spring and by 9.4\% relative to an end-to-end MDN.
In box pick-and-place, force adaptation maintains task success at 10/10 and increases both damage-free outcomes and reliability from 7/10 to 10/10.
The project page is available at \url{https://finger-force-estimation.pages.dev/}.
\end{abstract}

%
% Introduction
%
\section{Introduction}
\label{sec:introduction}

Box pick-and-place requires enough contact force to maintain a stable grasp without damaging the box.
This task is common in logistics and manufacturing, where robots handle packages with different physical properties~\cite{li2023package}.
Boxes with similar dimensions can differ in mass, contents, and structural strength.
These properties may be unknown before manipulation, so the same grasping motion can produce different outcomes across boxes.
Insufficient contact force can cause grasp or placement failure.
Excessive force can deform the box or damage its contents, leading to product loss and additional operating costs~\cite{aduh2024damage}.
Reliable handling therefore requires successful placement without box damage.

Force feedback can help maintain a stable grasp while limiting excessive interaction force~\cite{gabler2022force,ford2023gentle}.
Dedicated force and tactile sensors provide measurements at the contact surface~\cite{wang2024multitactile}.
However, integrating these sensors across multiple fingers requires additional hardware and calibration~\cite{wang2024multitactile}.
These requirements motivate estimating external forces from proprioceptive information already available from the robot~\cite{yigit2021position,oh2026factr2}.
External forces influence how the fingers respond to motion commands.
Proprioceptive measurements of this response may therefore provide useful information for estimating the forces acting on each finger.
This approach can provide per-finger force feedback without dedicated contact sensors during deployment.

The central challenge is to obtain per-finger force estimates that support successful box manipulation without damage.
When a position-controlled finger contacts a box, it may move less than commanded.
The difference between commanded and measured joint positions may provide a useful signal for estimating the contact force.
A virtual-spring model estimates force from displacement using a calibrated stiffness~\cite{erickson2003contact}.
However, unobserved contact locations can make the relationship between joint responses and interaction forces difficult to capture with a fixed analytical model.
This motivates learning a residual correction to the model-based force estimate from proprioceptive histories.

We hypothesize that per-finger force feedback improves manipulation reliability relative to open-loop finger commands.
To this end, we present \algonamelong{} (\algoname), an uncertainty-aware method that estimates per-finger external force magnitude by correcting a calibrated virtual-spring estimate with a residual distribution learned from proprioceptive histories.
We collect real-robot data pairing proprioceptive histories with reference force magnitudes measured by a force/torque (F/T) sensor to train the residual model.
The corrected force estimate provides continuous per-finger feedback without dedicated contact sensors during deployment.
The residual distribution also provides predictive uncertainty, which we examine as an empirical signal of ambiguity when force variations are weakly reflected in proprioceptive histories.
We evaluate force-estimation accuracy, the contribution of force adaptation to reliable box pick-and-place, and whether predictive uncertainty reflects this ambiguity.

The main contributions of this paper are twofold: 1) an uncertainty-aware residual method that estimates per-finger external force magnitude from proprioception for reliable box manipulation; and 2) a real-robot evaluation comparing accuracy against virtual-spring and end-to-end MDN baselines, assessing uncertainty as a signal of proprioceptive ambiguity, and testing whether force adaptation reduces box damage while maintaining task success relative to open-loop finger commands.
We used approximately 120 boxes across the real-robot manipulation experiments.

%
% Related work
%
\section{Related Work}
\label{sec:related_work}

Damage during robotic handling is a practical concern in warehouse logistics.
Package damage can result in product loss and higher operating costs~\cite{aduh2024damage}.
Prior work addresses manipulation damage before and during task execution.
Damage-prediction methods identify risky actions before execution~\cite{aduh2024damage}.
Force-sensitive grasping and gentle manipulation regulate physical interaction during execution to limit object damage~\cite{gabler2022force,ford2023gentle}.
Package-handling studies also examine box deformation and loss of grasp during physical transfer~\cite{tanaka2019box,li2023package}.
Force feedback is therefore relevant to completing the task while limiting box damage.

Force feedback can be obtained through direct sensing or proprioceptive estimation.
Dedicated force and tactile sensors provide direct measurements at the contact surface~\cite{wang2024multitactile}.
However, equipping multiple fingers with dedicated sensors increases hardware cost and requires calibration at each sensing location~\cite{oh2026factr2,yu2025calibration}.
Model-based methods estimate interaction without external contact sensors using robot dynamics, actuator measurements, or observed motion~\cite{colome2013external,yigit2021position,doan2024sensorless}.
A common model represents compliant interaction as a virtual spring that relates elastic displacement to force~\cite{erickson2003contact}.
The spring stiffness can be identified from measured interaction data, including through data-driven regression~\cite{uhlmann2023hybrid}.
A calibrated virtual spring provides the nominal force magnitude estimate in our residual formulation.

Data-driven force estimation can use direct regression or residual learning.
Direct methods estimate force or torque from synchronized robot measurements and reference labels~\cite{jung2022lstm,oh2026factr2,ballesteros2020proprioceptive}.
Hybrid methods retain model-based predictions by learning their residual errors~\cite{nguyentuong2010model,lim2025mobnet}.
Residual learning retains the nominal estimate while accounting for effects that are difficult to model.
Our formulation adds a proprioceptive residual to each finger's calibrated virtual-spring estimate.

The residual mapping can remain ambiguous when proprioceptive measurements do not fully describe the physical interaction.
A Mixture Density Network (MDN) represents the conditional residual distribution as a Gaussian mixture~\cite{bishop1994mdn}.
The conditional distribution admits a closed-form decomposition of predictive variance into within-component and between-component terms~\cite{choi2018uncertainty}.
We refer to these terms collectively as predictive uncertainty and analyze whether they reflect ambiguity in proprioceptive force estimation.

%
% Problem formulation
%
\section{Problem Formulation}
\label{sec:problem_formulation}

At deployment, the robot observes a history of proprioceptive signals.
Our proposed method uses this history to estimate per-finger external force magnitudes without dedicated force sensors.
The force estimates support continuous finger regulation, while predictive uncertainty is analyzed as an empirical descriptor of ambiguity in the residual prediction.

%
% Robot Configuration and Modeling Assumptions
%
\subsection{Robot Configuration and Modeling Assumptions}
\label{subsec:problem_configuration}

We consider a bimanual robot equipped with two position-controlled, five-fingered hands.
Each finger has four actuated joints.
The four non-thumb fingers have one abduction--adduction joint and three flexion joints, while the thumb uses a separate four-joint chain for opposition and flexion.
For finger $i\in\{1,\ldots,N\}$ at time $t$, the measured joint position is $\mathbf q_{i,t}\in\mathbb R^4$, and the commanded joint position is $\mathbf q^{\mathrm{cmd}}_{i,t}\in\mathbb R^4$.
Measured and commanded joint positions are updated at 100~Hz, matching the hand-control loop.
The main experiments use the proportional--derivative (PD) gain selected from three settings in Section~\ref{subsec:exp_gain}.

We adopt a virtual-spring model to obtain an initial force magnitude estimate for each finger.
Computing the spring displacement requires a defined point on the finger.
Exact contact location is difficult to determine from proprioceptive measurements alone.
We therefore assign one nominal contact point to the center of each fingertip surface.
For finger $i$, its current position $\mathbf p_i(\mathbf q_{i,t})$ is obtained from the measured joint position through forward kinematics.
We approximate the finger--object interaction as quasi-static when constructing the virtual-spring estimate.
The scope of this work is to estimate the corresponding external-force magnitude for each finger.

%
% Residual Force Magnitude Formulation
%
\subsection{Residual Force Magnitude Formulation}
\label{subsec:problem_force}

Under these assumptions, the displacement vector from the measured contact-point position to its commanded position provides a kinematic proxy for the loading direction.
Its norm defines the spring displacement.
For finger $i$, let $\delta_{i,t}=\|\mathbf p_i(\mathbf q^{\mathrm{cmd}}_{i,t})-\mathbf p_i(\mathbf q_{i,t})\|_2$ denote the spring displacement.
The virtual-spring force magnitude estimate is

\begin{equation}
f^{\mathrm{spr.}}_{i,t}
=
k_i \delta_{i,t},
\label{eq:virtual_spring}
\end{equation}
where $k_i\geq0$ is the calibrated virtual stiffness for finger $i$.
The superscript $\mathrm{spr.}$ denotes the virtual-spring estimate.
Section~\ref{subsec:method_prior} describes the stiffness calibration using a separate dataset.
The estimate represents an external-force magnitude in newtons under the assumed contact geometry.

The goal is to learn the residual between the reference force magnitude and the virtual-spring estimate from proprioceptive history.
Let $f^{\mathrm{ref.}}_{i,t}$ denote the reference force magnitude used during training and evaluation.
The residual target is $r_{i,t}=f^{\mathrm{ref.}}_{i,t}-f^{\mathrm{spr.}}_{i,t}$.
We model the relation between the proprioceptive history $\mathbf x_{i,t}$ and the force magnitude residual $r_{i,t}$ as the conditional distribution $p_\theta(r_{i,t}\mid\mathbf x_{i,t})$.
At deployment, $r_\theta(\mathbf x_{i,t})$ denotes the residual estimate obtained from this distribution.
The estimated external-force magnitude is

\begin{equation}
\hat f_{i,t}
=
f^{\mathrm{spr.}}_{i,t}
+
r_\theta(\mathbf x_{i,t}).
\label{eq:residual_force}
\end{equation}
In addition to the force estimate, we seek a predictive uncertainty signal that characterizes ambiguity in the residual prediction.
The force estimate is used for continuous per-finger regulation, while predictive uncertainty is analyzed as an empirical signal of force variations that are weakly reflected in the proprioceptive history.
Residual inference and predictive-uncertainty computation are described in Section~\ref{subsec:method_residual}.

\begin{figure*}[t]
  \centering
  {
    \includegraphics[width=\textwidth]{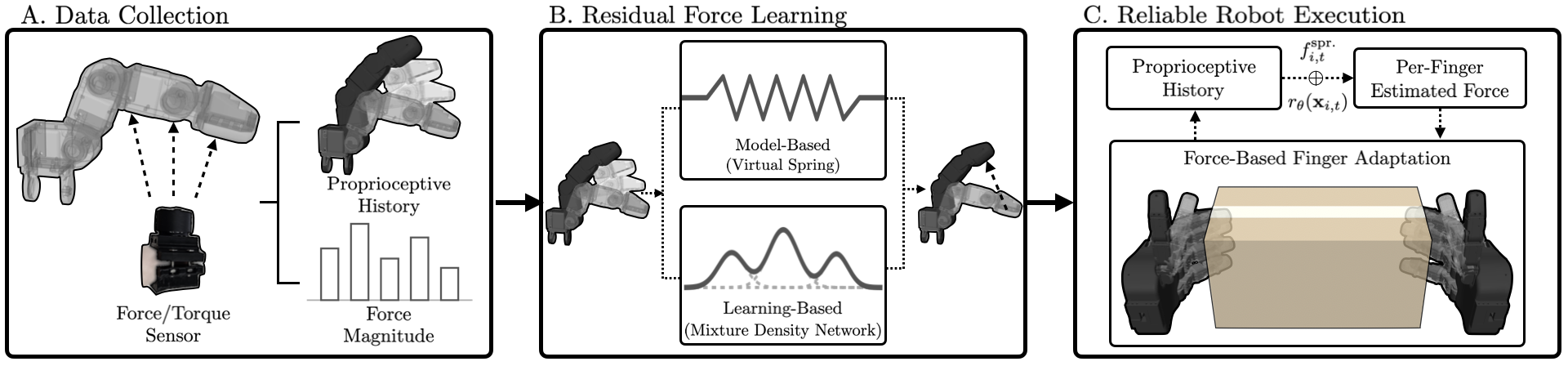}%
    \caption{
    Overview of \algoname{}.
    (A) A wearable F/T sensor collects reference force magnitudes synchronized with proprioceptive histories during finger motion.
    (B) The proprioceptive history $\mathbf{x}_{i,t}$ serves as input to \algoname{}.
    The calibrated virtual spring provides the model-based force estimate $f^{\mathrm{spr.}}_{i,t}$, while the MDN predicts the residual $r_\theta(\mathbf{x}_{i,t})$ and its predictive uncertainty.
    (C) The two force terms are added to obtain the per-finger estimate, $\hat f_{i,t}=f^{\mathrm{spr.}}_{i,t}+r_\theta(\mathbf{x}_{i,t})$.
    During deployment, this estimate provides feedback for force-based finger adaptation without dedicated contact sensors.}
    \label{fig:overview}
  }
\end{figure*}

%
% Reliable Box Manipulation
%
\subsection{Reliable Box Manipulation}
\label{subsec:problem_task}

We consider box pick-and-place as a representative handling operation in logistics.
The robot lifts a box placed in front of it and transfers the box to a designated shelf.
A trial is successful when the planned transfer and placement are completed.
A completed placement may still damage the box, whereas damage avoidance alone is insufficient if placement fails.
We therefore evaluate both outcomes.

We evaluate manipulation performance using trial-level reliability.
Let $S_n$ and $B_n$ denote task success and a damage-free box outcome in trial $n$, respectively.
The trial-level reliability indicator is defined as

\begin{equation}
\rho_n
=
\mathbb I\!\left[S_n \land B_n\right].
\label{eq:reliability}
\end{equation}
Thus, $\rho_n=1$ only when placement succeeds without box damage.
The task evaluation reports the proportion of reliable trials together with the separate success and damage rates.
Section~\ref{subsec:exp_task} specifies the seam-damage proxy.

%
% Proposed method
%
\section{Proposed Method}
\label{sec:method}

This section presents the proposed external-force magnitude estimation module and its use for box manipulation.
At each control step, \algoname{} estimates per-finger external force magnitudes from measured and commanded joint positions.
Each estimate combines a calibrated virtual-spring estimate with a residual predicted by the MDN.
Each finger then closes, holds, or opens according to whether its estimated force is below, within, or above a prescribed force band.
Fig.~\ref{fig:overview} provides an overview of data collection, per-finger force estimation, and force-based finger adaptation.
Sections~\ref{subsec:method_data} and~\ref{subsec:method_prior} describe training-data collection and virtual-spring calibration.
Section~\ref{subsec:method_residual} presents residual force estimation and the computation of predictive uncertainty.
Section~\ref{subsec:method_adaptation} describes how the estimated forces are used to update individual finger commands.

%
% Real Force-Data Collection
%
\subsection{Real Force-Data Collection}
\label{subsec:method_data}

We collect synchronized proprioceptive and force measurements from real-robot interactions to train and evaluate the force estimators.
Proprioceptive histories provide the estimator inputs, while a force/torque (F/T) sensor provides the reference force measurements.
The measured force magnitude $f^{\mathrm{ref.}}_{i,t}=\|\mathbf f^{\mathrm{f.t.}}_{i,t}\|_2$ serves as the reference force measurement during training and evaluation.
The F/T force direction and torque are not used for supervision.
The simulation--real force discrepancy reported in Section~\ref{subsec:exp_simreal} motivates using real measurements for training.
The F/T sensor is not used during box manipulation.

\begin{figure}[!t]
  \centering
  {
    \includegraphics[width=\columnwidth]{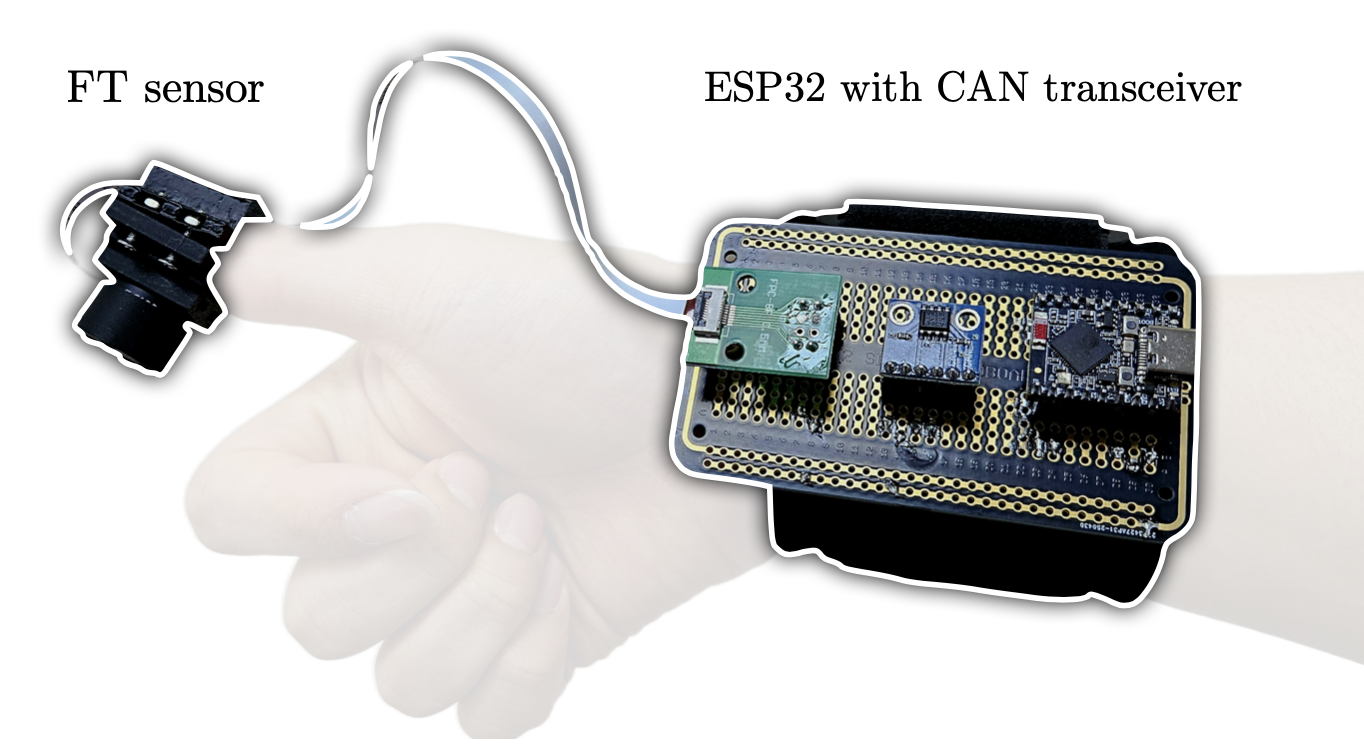}%
    \caption{
    Real force-data collection using a wearable F/T sensor.
    Contact interactions at multiple finger locations are synchronized with proprioceptive measurements.}
    \label{fig:data_collection}
  }
\end{figure}

A custom wearable device allows the sensing surface to be positioned at selected contact locations on the robot finger.
The device is built around an AFT20-D15 force/torque (F/T) sensor (AIDIN Robotics).
Fig.~\ref{fig:data_collection} shows the sensor attached to the operator's fingertip using a custom-designed fingertip mount.
An ESP32 microcontroller and an SN65HVD230 CAN transceiver are mounted on the wrist.
The microcontroller communicates with the F/T sensor over CAN through the transceiver.
The three-axis force vector $\mathbf f^{\mathrm{f.t.}}_{i,t}\in\mathbb R^3$ and the torque channels are transmitted to the host computer through a serial connection and recorded at 90~Hz.

During each collection sequence, the robot replays the same predefined motion for the selected finger.
The operator applies the instrumented surface at multiple locations on the finger while varying the contact force.
Contact location is varied during collection but is not provided to the estimator.
Data collection deliberately includes both contact interactions and free motion without external contact.
The collected force magnitude spans 0--20~N, and samples before, during, and after contact are retained.
The estimator dataset contains 25,000 synchronized samples per finger.

For each F/T measurement, the most recent $H=10$ robot-state samples up to the aligned timestamp form a 0.1-s proprioceptive history.
The corresponding reference force magnitude provides $f^{\mathrm{ref.}}_{i,t}$ and is paired with $f^{\mathrm{spr.}}_{i,t}$ to compute the residual label.
Measured and commanded joint positions are recorded at 100~Hz, while the F/T channels are recorded at 90~Hz.
The clocks of the two computers are synchronized before data collection, and the two streams are aligned by timestamp on the host computer.

%
% Virtual-Spring Calibration
%
\subsection{Virtual-Spring Calibration}
\label{subsec:method_prior}

We identify the stiffness in Eq.~\eqref{eq:virtual_spring} from a separate calibration dataset using least-squares fitting~\cite{erickson2003contact}.
This dataset contains approximately 5,000 synchronized samples per finger and does not overlap with the estimator dataset.
For each finger, the nonnegative stiffness $k_i$ is identified over its calibration samples:

\begin{equation}
k_i^*
=
\underset{k_i\geq0}{\arg\min}
\sum_j
\left(f^{\mathrm{ref.}}_{i,j}-k_i \delta_{i,j}\right)^2.
\label{eq:spring_calibration}
\end{equation}

The identified stiffness is fixed before residual-model training.
The nominal fingertip point defined in Section~\ref{subsec:problem_configuration} is used to compute the spring displacement $\delta$.
Together with the identified stiffness, it defines the virtual-spring estimate used to construct the residual labels.

%
% Residual Force Magnitude Estimation
%
\subsection{Residual Force Magnitude Estimation}
\label{subsec:method_residual}

The virtual spring estimates force magnitude from contact-induced tracking displacement using a calibrated stiffness~\cite{erickson2003contact}.
Its fixed stiffness and nominal contact point, however, cannot fully capture variations in contact geometry and finger compliance.
We therefore retain the virtual-spring estimate and learn only its residual error~\cite{nguyentuong2010model}.
This formulation anchors the prediction to the nominal physical model while using data-driven learning to account for unmodeled effects.

External contact can prevent a position-controlled finger from following its commanded motion.
The resulting deviation provides a signal related to the external load, while the measured joint position provides the configuration needed to interpret this signal~\cite{yigit2021position,oh2026factr2}.
For finger $i$, we define the joint tracking error as $\mathbf e_{i,t}=\mathbf q^{\mathrm{cmd}}_{i,t}-\mathbf q_{i,t}$.
A short history captures how the joint position and tracking error evolve during contact:

\begin{equation}
\mathbf x_{i,t}
=
\left[
\mathbf q_{i,t-H+1:t},
\mathbf e_{i,t-H+1:t}
\right].
\label{eq:proprioceptive_history}
\end{equation}
We use $H=10$, corresponding to a 0.1-s history at 100~Hz, without external force or tactile measurements at deployment.
Variations in unobserved contact conditions can produce different force residuals from similar proprioceptive histories.
To represent this conditional variability, we model $p_\theta(r\mid\mathbf x)$ using an MDN rather than assuming a single deterministic residual~\cite{bishop1994mdn,choi2018uncertainty}.

For $K$ Gaussian components, the residual distribution is

\begin{equation}
p_{\theta}(r\mid\mathbf x)
=
\sum_{k=1}^{K}
\pi_k(\mathbf x)
\mathcal N\!\left(
r;\mu_k(\mathbf x),\sigma_k^2(\mathbf x)
\right),
\label{eq:residual_mdn}
\end{equation}
where $\pi_k$, $\mu_k$, and $\sigma_k$ denote the weight, mean, and standard deviation of component $k$, respectively.
Each component represents a possible range of the residual force magnitude, and its weight describes the relative likelihood of that range.
The network parameters are learned by minimizing the negative log likelihood of the residual labels:

\begin{equation}
\mathcal L_{\mathrm{res}}(\theta)
=
-\sum_{(\mathbf x,r)\in\mathcal D_{\mathrm{train}}}
\log p_{\theta}(r\mid\mathbf x).
\label{eq:residual_loss}
\end{equation}
This objective encourages the predicted distribution to assign high likelihood to the residual observed for each proprioceptive history.
At inference, the mean of the highest-weight component provides the residual estimate:

\begin{equation}
r_\theta(\mathbf x)
=
\mu_{\arg\max_k\pi_k(\mathbf x)}(\mathbf x).
\label{eq:residual_inference}
\end{equation}
The force magnitude estimate follows Eq.~\eqref{eq:residual_force} and is clipped at zero to maintain a nonnegative value.

An additional property of the residual MDN is that its predictive variance has a closed-form decomposition by the law of total variance~\cite{choi2018uncertainty}.
Let $\bar r(\mathbf x)=\sum_k\pi_k(\mathbf x)\mu_k(\mathbf x)$ denote the mixture mean.
The predictive variance of the residual decomposes as

\begin{equation}
\begin{aligned}
\mathbb V_\theta(r\mid\mathbf x)
&=
\mathbb E_{k\sim\pi}\!\left[\mathbb V(r\mid\mathbf x,k)\right]
+
\mathbb V_{k\sim\pi}\!\left(\mathbb E[r\mid\mathbf x,k]\right)
\\
&=
\underbrace{\sum_{k=1}^{K}\pi_k(\mathbf x)\sigma_k^2(\mathbf x)}_{
u_\theta^{\mathrm{alea.}}(\mathbf x)}
+
\underbrace{\sum_{k=1}^{K}\pi_k(\mathbf x)
\left(\mu_k(\mathbf x)-\bar r(\mathbf x)\right)^2}_{
u_\theta^{\mathrm{epis.}}(\mathbf x)}.
\end{aligned}
\label{eq:residual_uncertainty}
\end{equation}
Following~\cite{choi2018uncertainty}, the first term, $u_\theta^{\mathrm{alea.}}$, is the within-component variance and is interpreted as aleatoric uncertainty.
The second term, $u_\theta^{\mathrm{epis.}}$, is the between-component variance and is interpreted as epistemic uncertainty.
Together, these terms constitute the predictive variance of the residual MDN and are obtained from the same forward pass without sampling.
We refer to them collectively as predictive uncertainty in the remainder of the paper and evaluate them as empirical signals rather than calibrated measures of model uncertainty.
Section~\ref{subsec:exp_uncertainty} examines whether predictive uncertainty reflects ambiguity in proprioceptive force estimation.

%
% Force-Based Finger Adaptation
%
\subsection{Force-Based Finger Adaptation}
\label{subsec:method_adaptation}

We assume that the object contacts the inner surfaces of the fingers during grasping.
The nominal finger motion closes each finger from the measured initial configuration toward a fixed target hand configuration~\cite{park2026grit,lin2025typetele}.
We represent this movement by a normalized motion progress from zero to one.
Our objective is to adapt each finger command to keep its estimated force magnitude near a prescribed level for stable grasping without excessive contact force.

Let $\Delta\mathbf q_{i,t}$ denote the joint-position increment toward the target configuration, and let $f_i^{\mathrm{th.}}$ denote the force magnitude threshold.
A $\pm1$-N deadband around this threshold reduces command changes caused by small estimation fluctuations.
The coefficient $\alpha_{i,t}$ takes values of $1$, $0$, and $-1$ to close, hold, and open the finger, respectively.
Here, $\mathcal Q_i$ denotes the admissible joint-position range, and $\Pi_{\mathcal Q_i}$ projects the command into this range.
The command is updated as

\begin{equation}
\mathbf q^{\mathrm{cmd}}_{i,t+1}
=
\Pi_{\mathcal Q_i}\!\left(
\mathbf q^{\mathrm{cmd}}_{i,t}
+
\alpha_{i,t}
\Delta\mathbf q_{i,t}
\right).
\label{eq:force_adaptation}
\end{equation}
Each finger closes below the deadband, holds within it, and moves in the opening direction above it.
This force-based regulation adapts each finger command independently according to its estimated force magnitude.
Fingers that reach contact earlier can hold or open while the others continue closing.

%
% Experiments
%
\section{Experiments}
\label{sec:experiments}

This section evaluates the accuracy of the proposed force estimation module and the effect of force adaptation on box manipulation.
Section~\ref{subsec:exp_setup} describes the experimental setup.
The simulation--real comparison reveals different force responses under matched finger commands, motivating training with real force measurements (Section~\ref{subsec:exp_simreal}).
On the real dataset, residual learning reduces mean force-estimation error relative to the virtual spring and the end-to-end MDN (Section~\ref{subsec:exp_estimation}).

Using these force estimates for finger adaptation reduces box damage while maintaining task success in the evaluated pick-and-place motion (Section~\ref{subsec:exp_task}).
However, estimation accuracy alone does not determine manipulation performance: the lowest controller-gain setting yields the most accurate estimates but the lowest task success (Section~\ref{subsec:exp_gain}).
The analysis also considers loading conditions in which force variations are weakly reflected in proprioception.
Mixture-component disagreement increases under these conditions, providing an empirical signal of ambiguity (Section~\ref{subsec:exp_uncertainty}).
Section~\ref{subsec:exp_limitations} discusses the limitations of these findings.

%
% Experimental Setup
%
\subsection{Experimental Setup}
\label{subsec:exp_setup}

\begin{figure}[t]
  \centering
  \includegraphics[width=\columnwidth]{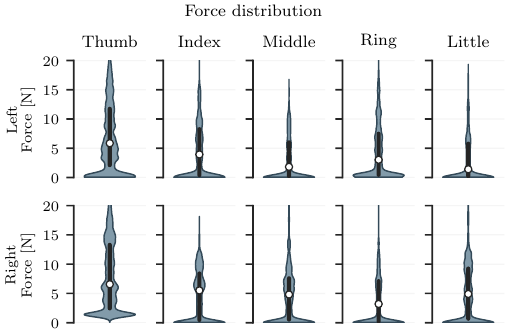}
  \caption{Force distributions in the collected dataset for ten fingers.
  The top and bottom rows show the distributions of force data collected from the left and right hands, respectively.
  Within each finger, wider regions indicate more samples at that force level.
  White circles mark the median force, and black bars span the middle 50\% of samples.}
  \label{fig:dataset_distribution}
\end{figure}

Experiments use the bimanual robot and five-fingered hands described in Section~\ref{subsec:problem_configuration}.
Force data are collected with the wearable F/T tool in Section~\ref{subsec:method_data}.
The main estimator dataset contains 25,000 synchronized samples per finger over 0--20~N at the selected proportional--derivative (PD) gain setting $(K_p,K_d)=(200,400)$.
Its force distributions are shown in Fig.~\ref{fig:dataset_distribution}.
A separate virtual-spring calibration set contains approximately 5,000 synchronized samples per finger.
Robot states are sampled at 100~Hz, and the F/T measurements are sampled at 90~Hz.
The two streams are aligned by timestamp on the host computer.
The three-axis force magnitude provides the reference value.

We train one estimator per finger from a 0.1-s proprioceptive history of $H=10$ robot observations.
Each observation contains four measured joint positions and four tracking errors, giving an 80-dimensional input.
All learned estimators use two 64-unit fully connected hidden layers with tanh activations.
Each MDN has $K=10$ Gaussian components and 11{,}294 trainable parameters.
Training uses AdamW for 1{,}000 epochs with a batch size of 512.

Evaluation uses row-level five-fold cross-validation stratified by force magnitude bins.
Each fold contains 20,000 training samples and 5,000 test samples per finger, with identical splits across the learned methods.
Input standardization is computed only from the training samples in each fold.
We report test root-mean-square error (RMSE) in newtons as the mean $\pm$ sample standard deviation across five folds.
For the manipulation experiments, the residual MDN is retrained on the complete estimator dataset using the same architecture and hyperparameter settings.
Force magnitude inference runs at 100~Hz, matching the hand-control loop.

\begin{table*}[t]
  \centering
  \caption{
  Comparison of per-finger external force magnitude estimation accuracy using row-level five-fold cross-validation stratified by force magnitude bins.}
  \label{tab:force_estimation}
  \small
  \begin{tabular*}{\textwidth}{@{\extracolsep{\fill}}llccc@{}}
  \toprule
  \multirow{2}{*}{Hand} & \multirow{2}{*}{Finger}
  & Virtual spring~\cite{erickson2003contact}
  & End-to-end MDN~\cite{gao2020compliance}
  & Residual MDN (Ours) \\
  \cmidrule(lr){3-5}
  & & \multicolumn{3}{c}{RMSE (N) $\downarrow$} \\
  \midrule
  \multirow{6}{*}{Left}
  & Thumb  & $5.327 \pm 0.012$ & $0.790 \pm 0.029$ & \textbf{0.723 $\pm$ 0.024} \\
  & Index  & $3.203 \pm 0.042$ & $0.192 \pm 0.030$ & \textbf{0.178 $\pm$ 0.044} \\
  & Middle & $2.545 \pm 0.026$ & \textbf{0.093 $\pm$ 0.002} & $0.096 \pm 0.004$ \\
  & Ring   & $3.467 \pm 0.030$ & $0.140 \pm 0.059$ & \textbf{0.119 $\pm$ 0.021} \\
  & Little & $2.135 \pm 0.033$ & \textbf{0.103 $\pm$ 0.004} & $0.106 \pm 0.006$ \\
  \cmidrule(lr){2-5}
  & \textit{Mean} & $3.335 \pm 0.011$ & $0.264 \pm 0.016$ & \textbf{0.245 $\pm$ 0.008} \\
  \midrule
  \multirow{6}{*}{Right}
  & Thumb  & $5.361 \pm 0.034$ & $0.718 \pm 0.052$ & \textbf{0.669 $\pm$ 0.056} \\
  & Index  & $3.092 \pm 0.033$ & $0.207 \pm 0.014$ & \textbf{0.161 $\pm$ 0.011} \\
  & Middle & $3.472 \pm 0.047$ & \textbf{0.132 $\pm$ 0.019} & $0.147 \pm 0.027$ \\
  & Ring   & $3.075 \pm 0.035$ & $0.184 \pm 0.087$ & \textbf{0.123 $\pm$ 0.008} \\
  & Little & $3.254 \pm 0.040$ & $0.214 \pm 0.018$ & \textbf{0.186 $\pm$ 0.025} \\
  \cmidrule(lr){2-5}
  & \textit{Mean} & $3.651 \pm 0.014$ & $0.291 \pm 0.016$ & \textbf{0.257 $\pm$ 0.016} \\
  \midrule
  Overall & \textit{Mean} & $3.493 \pm 0.007$ & $0.277 \pm 0.014$ & \textbf{0.251 $\pm$ 0.010} \\
  \bottomrule
  \end{tabular*}
\end{table*}

For the predictive-uncertainty evaluation, we additionally collect 3,000 synchronized samples per finger under a loading condition designed to produce substantial force variation with limited changes in proprioceptive signals.
A vertical load is applied at the distal end of each finger while the commanded finger configuration is held constant.
The same F/T measurement setup provides the corresponding reference force magnitudes.

The manipulation experiments use a cardboard box measuring $395\times275\times240$~mm (width $\times$ depth $\times$ height), with a mass of 2.0~kg, and a designated shelf as the placement target.
A trial is considered successful when the box remains on the shelf at the end of the predefined pick-and-place motion.
Tape separation from either side of the designated seam is recorded as box damage.
Each box is prepared with 4.5~cm of low-adhesion packaging tape adhered to either side of the seam, and a new box is used for every trial.
Reliability follows Eq.~\eqref{eq:reliability} and counts trials that complete the task without recorded damage.
Within each comparison, the initial box pose, box specification, and controller parameters other than the evaluated component remain fixed.
External force and tactile sensors are not used during manipulation.
We use a 6-N threshold for all fingers, selected in preliminary grasp trials and fixed for task evaluation.

\begin{figure}[t]
  \centering
    {\includegraphics[width=0.9\columnwidth]{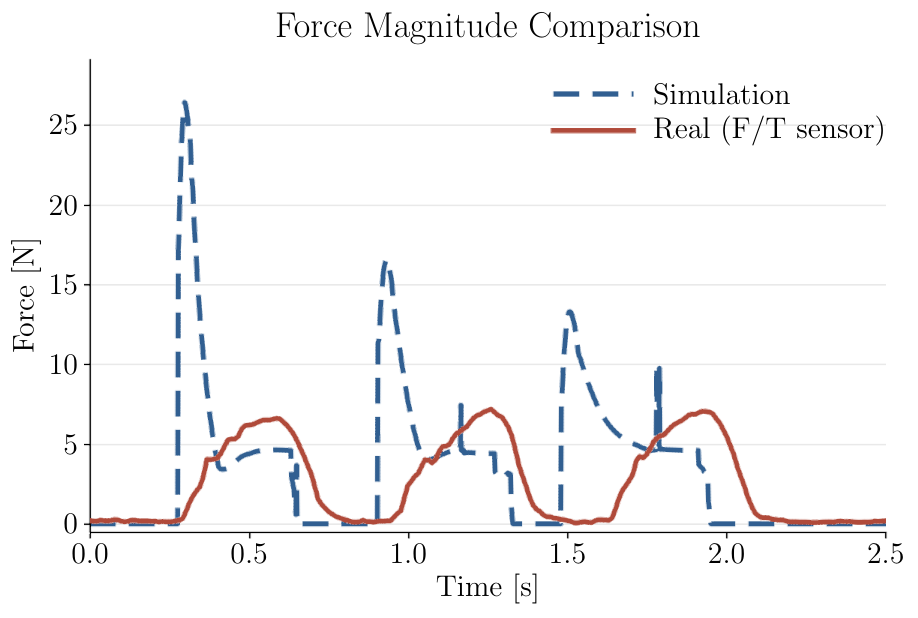}%
    \caption{
    Comparison of force magnitude traces from matched tapping motions in MuJoCo and on the real robot.}
    \label{fig:sim_real}
  }
\end{figure}

%
% Simulation--Real Force Magnitude Comparison
%
\subsection{Simulation--Real Force Magnitude Comparison}
\label{subsec:exp_simreal}

This experiment examines how closely simulated force responses match physical measurements under the same finger commands.
The simulated and measured traces differ in profile, with an RMSE of 5.287~N (Fig.~\ref{fig:sim_real}).
Contact forces depend not only on the commanded motion but also on the interaction between the finger and the contact surface.
The observed discrepancy motivates using real F/T measurements to train the physical-robot force estimators.

For this comparison, the left index finger executes the same three-tap command sequence in MuJoCo and on the physical robot~\cite{todorov2012mujoco}.
The commanded trajectory, initial configuration, PD gains, and effort limit are matched across the two setups.
The physical experiment uses an F/T-instrumented flat surface, while the simulation uses its nominal contact geometry.
Differences in contact modeling, hardware compliance, and actuator response may contribute to the observed force discrepancy~\cite{lelidec2024contact}.
The comparison is limited to one finger and one three-tap sequence.

%
% Force Magnitude Estimation
%
\subsection{Force Magnitude Estimation}
\label{subsec:exp_estimation}

This experiment evaluates whether learning a residual correction improves per-finger force magnitude estimation.
The residual MDN achieves the lowest mean RMSE among the three evaluated estimators.
Across ten fingers, its mean RMSE is 0.251~N, compared with 3.493~N for the virtual spring and 0.277~N for the end-to-end MDN (Table~\ref{tab:force_estimation}).
The same trend appears in the averages of both hands.
These results support combining a calibrated force model with a learned residual correction under the evaluated conditions.

We compare all three estimators on identical cross-validation folds of the real contact dataset.
The virtual spring~\cite{erickson2003contact} estimates force using calibrated stiffness and displacement at a fixed contact point.
The end-to-end MDN predicts the reference force magnitude directly from proprioceptive histories, following prior MDN-based force prediction~\cite{gao2020compliance}.
The residual MDN instead predicts the error of the virtual-spring estimate from the same histories.
Both MDNs use the same network architecture.
This comparison isolates the residual formulation while keeping the learned models' inputs, architecture, and evaluation samples consistent.

We also evaluate whether predicting a residual distribution improves accuracy over predicting a single residual value.
The residual MDN achieves lower mean RMSE than the residual MLP, reducing the overall error from 0.270~N to 0.251~N (Table~\ref{tab:estimator_ablation}).
The residual MLP differs only in replacing the MDN output with a scalar prediction.
The reduction appears in the five-finger averages of both hands.
This result may reflect the mixture model's ability to represent alternative residual values when different contact conditions produce similar proprioceptive histories.

\begin{table}[t]
  \centering
  \caption{
  Ablation of residual output models for external force magnitude estimation.}
  \label{tab:estimator_ablation}
  \small
  \renewcommand{\arraystretch}{1.05}
  \begin{tabular*}{\columnwidth}{@{\extracolsep{\fill}}lcc@{}}
  \toprule
  \multirow{2}{*}{Hand} & Residual MLP & Residual MDN (Ours) \\
  \cmidrule(lr){2-3}
  & \multicolumn{2}{c}{RMSE (N) $\downarrow$} \\
  \midrule
  Left & $0.261 \pm 0.004$ & \textbf{0.245 $\pm$ 0.008} \\
  Right & $0.280 \pm 0.004$ & \textbf{0.257 $\pm$ 0.016} \\
  \midrule
  Overall & $0.270 \pm 0.004$ & \textbf{0.251 $\pm$ 0.010} \\
  \bottomrule
\end{tabular*}
\end{table}

\begin{figure*}[!t]
  \centering
  {%
    \includegraphics[width=\textwidth]{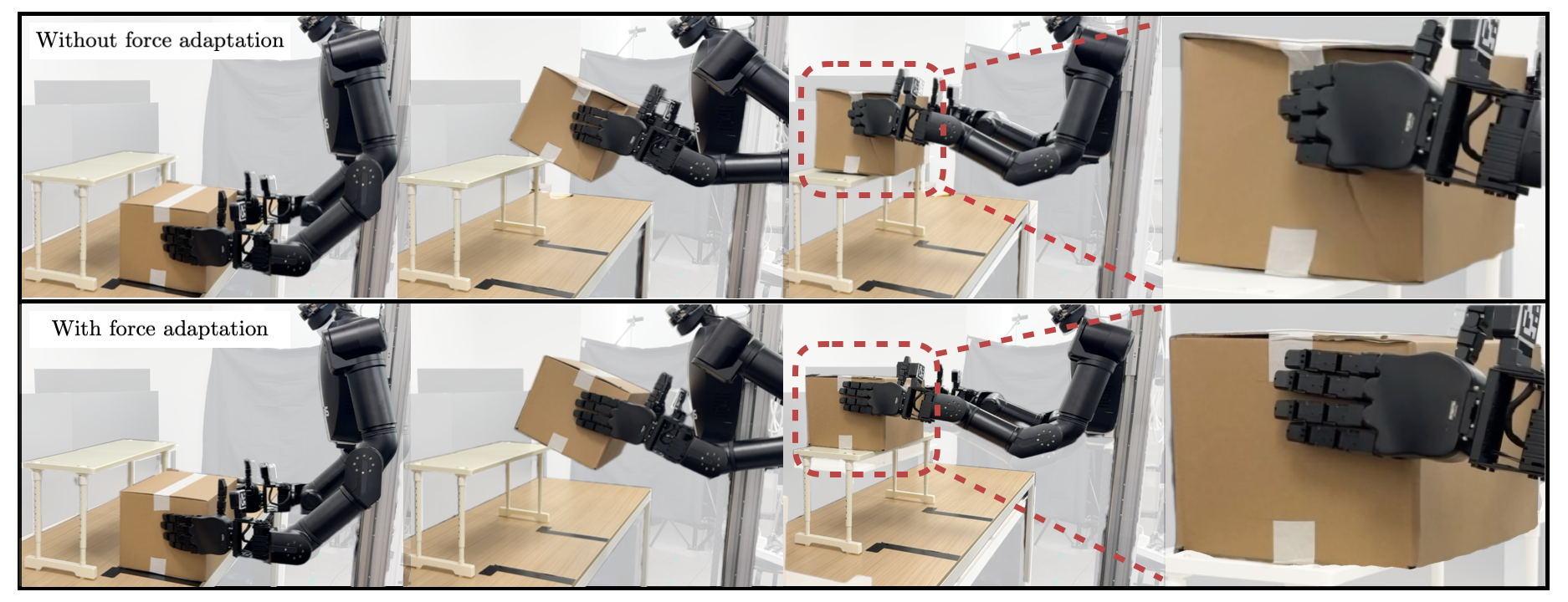}%
    \caption{
    Snapshots of real-robot box pick-and-place with and without force-based finger adaptation.
    The enlarged views show box deformation and damage at the designated seam without adaptation, whereas the seam remains intact with adaptation.}
  \label{fig:real_robot}
  }
\end{figure*}

%
% Reliable Box Pick-and-Place
%
\subsection{Reliable Box Pick-and-Place}
\label{subsec:exp_task}

This experiment evaluates whether per-finger force adaptation improves reliability in box pick-and-place.
Force adaptation reduces box damage while maintaining task success in the evaluated motion.
Both conditions complete 10/10 trials, while damage-free outcomes increase from 7/10 with fixed finger commands to 10/10 with adaptation (Table~\ref{tab:pick_place}).
Reliability consequently increases from 7/10 to 10/10.
The improvement comes from avoiding damage during successful placement, rather than increasing the task-completion rate.

We compare fixed finger commands with force adaptation driven by the residual MDN estimates.
The baseline executes the predefined finger commands without adaptation.
The comparison adjusts individual finger commands using the estimated forces, as described in Section~\ref{subsec:method_adaptation}.
Both conditions use the same arm motion, target hand configuration, placement target, and initial box pose.
Each condition is evaluated over 10 trials, with a new box used for every trial.
Task success and seam damage are recorded separately using the criteria in Section~\ref{subsec:exp_setup}.

Fig.~\ref{fig:real_robot} illustrates the corresponding difference in box deformation and seam damage.
With fixed finger commands, fingers continue closing after contact, which may increase local loading on the box.
Force adaptation allows individual fingers to hold or open according to their estimated force while the remaining fingers continue closing.
The observed reduction in seam damage is consistent with limiting excessive loading through this independent regulation.
Under the evaluated box specification and motion, force adaptation improves reliability by reducing damage while preserving task success.

\begin{table}[t]
  \centering
  \caption{
  Comparison of force adaptation under one predefined box pick-and-place motion over 10 trials per condition.}
  \label{tab:pick_place}
  \small
  \setlength{\tabcolsep}{3pt}
  \renewcommand{\arraystretch}{1.05}
  \begin{tabular}{
    >{\raggedright\arraybackslash}m{0.455\columnwidth}
    *{3}{>{\centering\arraybackslash}m{0.15\columnwidth}}}
  \toprule
  Method & Success $\uparrow$
  & \shortstack{Damage-\\free $\uparrow$}
  & Reliability $\uparrow$ \\
  \midrule
  No force adaptation & \textbf{10/10} & $7/10$ & $7/10$ \\
  Force adaptation (Ours) & \textbf{10/10} & \textbf{10/10} & \textbf{10/10} \\
  \bottomrule
  \end{tabular}
  \vspace{3mm}
\end{table}

%
% Controller-Gain Sensitivity
%
\subsection{Controller-Gain Sensitivity}
\label{subsec:exp_gain}

The finger PD gain influences both the proprioceptive signal used for force estimation and the physical response of the force-adaptive controller.
With $(K_p,K_d)=(100,200)$, external contact is expected to produce larger tracking deviations, which may make force magnitude easier to estimate, but weaker command tracking may limit grasp execution.
With $(K_p,K_d)=(300,600)$, stronger command tracking is expected to reduce those deviations and can increase local loading during contact even as the finger commands are adapted.
We therefore hypothesize a trade-off: the setting that minimizes force-estimation error does not necessarily maximize manipulation reliability under force adaptation.

We evaluate three PD-gain pairs.
The tested $(K_p,K_d)$ values are $(100,200)$, $(200,400)$, and $(300,600)$.
For each setting, the estimator is trained and evaluated on 25,000 force samples per finger collected at the corresponding gain.
During manipulation, each gain setting uses its corresponding residual MDN for the force-based finger adaptation in Section~\ref{subsec:method_adaptation}.
All settings use the same force threshold, adaptation rule, box, and predefined motion for 10 trials each.

Table~\ref{tab:gain_sensitivity} summarizes the estimation and manipulation results across the three gain settings.
The $(100,200)$ setting achieves the lowest RMSE but produces the lowest task-success rate.
This result suggests that larger contact-induced tracking errors provide a clearer signal for force estimation, while weaker position tracking limits task execution.
The $(300,600)$ setting achieves task success in 9/10 trials and damage-free outcomes in 8/10 trials.
Interestingly, we observed that the robot could not form a stable grasp because the high gains made the fingers too stiff.
The $(200,400)$ setting balances these effects and achieves the highest reliability of 10/10.
This comparison shows that the controller gain should be selected jointly for proprioceptive force estimation and force-adaptive manipulation.

\begin{table}[t]
  \centering
  \caption{
  Controller-gain sensitivity of force magnitude estimation and force-adaptive box manipulation.}
  \label{tab:gain_sensitivity}
  \small
  \setlength{\tabcolsep}{2.5pt}
  \begin{tabular*}{\columnwidth}{@{\extracolsep{\fill}}lcccc@{}}
  \toprule
  & \multirow{2}{*}{RMSE (N) $\downarrow$}
  & \multicolumn{3}{c}{Box manipulation (/10)} \\
  \cmidrule(lr){3-5}
  $(K_p,K_d)$ & & Success $\uparrow$ & \shortstack{Damage-\\free $\uparrow$}
  & Reliability $\uparrow$ \\
  \midrule
  $(100,200)$ & \textbf{0.233 $\pm$ 0.005} & 4/10 & \textbf{10/10} & 4/10 \\
  $(200,400)$ & $0.251 \pm 0.010$ & \textbf{10/10} & \textbf{10/10} & \textbf{10/10} \\
  $(300,600)$ & $0.255 \pm 0.008$ & 9/10 & 8/10 & 7/10 \\
  \bottomrule
\end{tabular*}
\end{table}

%
% Predictive Uncertainty Under Proprioceptive Ambiguity
%
\subsection{Predictive Uncertainty Under Proprioceptive Ambiguity}
\label{subsec:exp_uncertainty}

Proprioception enables force feedback without dedicated contact sensors, but external loading is reflected only through changes in joint position and tracking error.
Consider an interaction in which the finger structure bears a vertical load while the commanded finger configuration remains fixed.
The applied force can vary substantially with little change in the proprioceptive history.
Our proposed method (FEEL) uses an MDN to represent this ambiguity as distinct force-residual modes conditioned on a given history.
This distribution can express alternative residual outcomes, but it cannot resolve interaction information absent from the proprioceptive input.
Recognizing this unresolved ambiguity is important when interpreting the resulting force estimate.
We therefore hypothesize that mixture-component disagreement can provide an empirical signal of such cases.

To test this hypothesis, we compare Force through joints and Force through structure.
Force through joints uses held-out test samples collected with the same predefined finger-motion protocol as the training data.
Force through structure uses 3,000 separately collected samples per finger across all ten fingers.
These samples are obtained by applying a vertical load at the distal end of each finger while its commanded configuration remains fixed.
The samples have a mean reference force magnitude of $10.909 \pm 6.030$~N across both hands and are not used for training or input standardization.
Both conditions are evaluated using the same per-finger checkpoints and fold-specific standardization statistics.
Table~\ref{tab:uncertainty_loading} reports the mean $\pm$ sample standard deviation across five folds after averaging each uncertainty term over samples and fingers.

Mean aleatoric uncertainty remains similar across the two conditions, whereas mean epistemic uncertainty is higher for Force through structure.
This contrast indicates greater separation among the force-residual modes rather than greater variance within each mode.
Component disagreement can therefore be interpreted as a practical proxy for interactions in which force is not uniquely determined by proprioception.
It signals unresolved ambiguity rather than recovering the missing force information.
Future work may address this limitation by combining proprioception with complementary contact-sensitive measurements to capture force variations that are weakly reflected in joint responses.

%
% Limitations and Future Work
%
\subsection{Limitations and Future Work}
\label{subsec:exp_limitations}

The virtual spring assumes a single nominal fingertip point, and the residual model predicts one force magnitude per finger.
This formulation cannot localize contact, recover force direction, or resolve simultaneous contacts, and its accuracy decreases when different force conditions produce similar proprioceptive histories.
Training requires per-finger stiffness calibration and synchronized F/T supervision, although deployment does not require contact sensors.
The estimator dataset was collected as continuous interaction streams rather than independent trials, and we therefore use row-level stratified cross-validation.
As a result, temporally adjacent history windows may appear in different folds, while generalization across independently collected sequences remains to be evaluated.
Manipulation was evaluated over 10 trials per condition using one box specification and one motion, while the simulation--real comparison covers one finger and one three-tap sequence.
Future work will investigate richer contact representations, fuse complementary contact-sensitive measurements with proprioception, use calibrated simulation to reduce real-data collection, and evaluate additional objects and tasks.

\begin{table}[t]
  \centering
  \caption{
  Comparison of predictive uncertainty for Force through joints and Force through structure.}
  \label{tab:uncertainty_loading}
  \small
  \setlength{\tabcolsep}{6pt}
  \renewcommand{\arraystretch}{1.05}
  \begin{tabular*}{\columnwidth}{@{\extracolsep{\fill}}lcc@{}}
  \toprule
  Interaction condition
  & \begin{tabular}[c]{@{}c@{}}
      Aleatoric\\uncertainty
    \end{tabular}
  & \begin{tabular}[c]{@{}c@{}}
      Epistemic\\uncertainty
    \end{tabular} \\
  \midrule
  Force through joints
  & \textbf{0.370 $\pm$ 0.001}
  & 0.102 $\pm$ 0.005 \\
  Force through structure
  & 0.364 $\pm$ 0.005
  & \textbf{0.172 $\pm$ 0.088} \\
  \bottomrule
  \end{tabular*}
\end{table}

%
% Conclusion
%
\section{Conclusion}
\label{sec:conclusion}

We presented \algoname, an uncertainty-aware residual method for estimating per-finger external force magnitude from proprioception.
\algoname{} improved force-estimation accuracy over the evaluated baselines and enabled force adaptation that reduced box damage while maintaining task success.
Mixture-component disagreement provided an empirical signal when force variations were weakly reflected in proprioception.
These results establish \algoname{} as a practical approach to reliable box manipulation without dedicated contact sensors during deployment.

%
% References
%

\end{document}